\documentclass[conference]{IEEEtran}
\usepackage{times}
\usepackage{booktabs}
\usepackage{multirow}
\usepackage{graphicx}
\usepackage{amsmath}
\usepackage{amssymb}
\usepackage{algorithm}
\usepackage{algpseudocode}
\usepackage{gensymb}
\usepackage{xcolor}
\usepackage{tablefootnote}
\usepackage{twemojis}
\let\labelindent\relax
\usepackage{soul}
\usepackage{makecell}

\usepackage[numbers]{natbib}
\usepackage{multicol}
\usepackage[bookmarks=true]{hyperref}
\usepackage[inline]{enumitem}

\newcommand{\methodname}{\textsc{ScrewMimic}}

\begin{document}

\title{\texttwemoji{nut and bolt}\methodname{}: Bimanual Imitation from\\Human Videos with Screw Space Projection}

\author{Arpit Bahety, Priyanka Mandikal, Ben Abbatematteo, Roberto Martín-Martín \\
The University of Texas at Austin}



%

\maketitle

\begin{abstract}
Bimanual manipulation is a longstanding challenge in robotics due to the large number of degrees of freedom and the strict spatial and temporal synchronization required to generate meaningful behavior. 
Humans learn bimanual manipulation skills by watching other humans and by refining their abilities through play. 
In this work, we aim to enable robots to learn bimanual manipulation behaviors from human video demonstrations and fine-tune them through interaction. 
Inspired by seminal work in psychology and biomechanics, we propose modeling the interaction between two hands as a serial kinematic linkage --- as a screw motion, in particular, that we use to define a new action space for bimanual manipulation: screw actions. 
We introduce \methodname{}, a framework that leverages this novel action representation to facilitate learning from human demonstration and self-supervised policy fine-tuning. 
Our experiments demonstrate that \methodname{} is able to learn several complex bimanual behaviors from a single human video demonstration, and that it outperforms baselines that interpret demonstrations and fine-tune directly in the original space of motion of both arms. For more information and video results, \url{https://robin-lab.cs.utexas.edu/ScrewMimic/}
\end{abstract}

\IEEEpeerreviewmaketitle

\section{Introduction}
\label{s_intro}


Manipulation in human environments often requires coordinating the motion of two arms, e.g., opening a bottle, cutting a block in two pieces, or stirring a pot. 
In \textit{dexterous bimanual manipulation}, the agent has to generate behavior for both arms that are synchronized spatially and temporally, rendering it even more complex to generate than two independent unimanual manipulations.
Due to its complexity, in nature, this kind of behavior is almost unique to higher-level primates~\cite{heldstab2016manipulation,Kazennikov1999NeuralAO,Donchin1998PrimaryMC,Kelso1984PhaseTA}, and it requires several years to fully develop in humans~\cite{adolph1997learning,barral2006developmental}, being mastered only after a significant amount of time of observing expert bimanual agents and practicing through trial-and-error.
This work aims to endow robots with novel capabilities to learn bimanual manipulation tasks.

Learning to generate dexterous bimanual manipulation in robots is challenging due to the large state and action spaces resulting from the two arms, and the strict requirements of spatial and temporal synchronization between them to achieve success~\cite{koga1992experiments,smith2012dual,Colom2018DimensionalityRF}. As a result, exploring randomly in this space is prohibitively difficult, especially on real robot hardware, limiting some of the successes to simulation~\cite{Kroemer2015TowardsLH,Luck2017ExtractingBS,chen2022towards,Kataoka2022BiManualMA}. A promising approach to reduce the challenge of searching for a successful bimanual manipulation policy is to observe a human performing a bimanual manipulation and imitate it. However, due to the morphology differences between the human and the robot, the direct execution of the observed bimanual interaction may not be successful, necessitating an exploratory refinement to adapt to the robot's embodiment and capabilities, which reintroduces the challenges of exploring directly in the space of motion of both arms.

\begin{figure}[t]
    \centering
    \includegraphics[width=1\columnwidth]{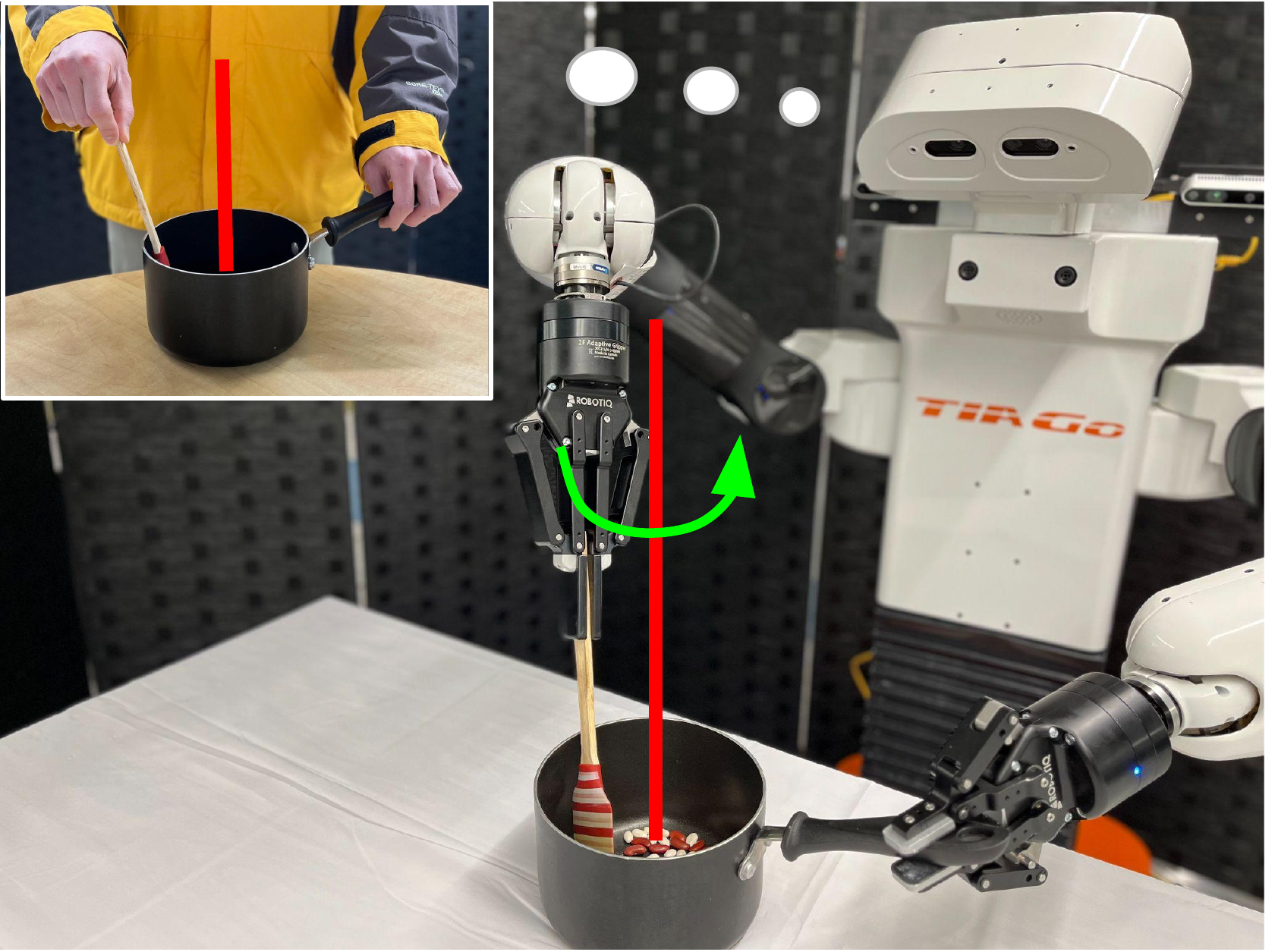}
    \caption{\textbf{Bimanual manipulation} tasks can be represented by a screw axis (red line) constraining and synchronizing the motion of both hands. \textbf{\methodname} maps a single human demonstration into a screw axis, improves it with an iterative interactive exploration procedure, and learns to predict it for new object instances and poses, enabling their manipulation.}
    \label{fig:pullfig} 
    \vspace*{-2em}
\end{figure}

The main insight in this work is that for many bimanual manipulation tasks, the relative motion between hands can be explained by a simple one-degree-of-freedom (1-DoF) screw joint. This virtual joint constrains the motion in a way that matches an existing physical constraint in the environment (e.g., when opening a laptop or a bottle with both hands) or just facilitates the manipulation (e.g., when cutting a block or stirring a pot, see Fig.~\ref{fig:pullfig}). The type of 1-DoF screw joint ---prismatic, revolute, screw--- captures different modalities of bimanual manipulation, while the screw joint parameters fully specify the motion. 
This insight works at several levels: in perception, it serves as a prior for interpreting noisy sensor signals and facilitates understanding a human-demonstrated bimanual manipulation.
And, in exploration, it provides an action space where both arm motions are coordinated by design, allowing efficient action fine-tuning to find successful behaviors with the real robot's embodiment.

We present a novel method, \methodname{}, that leverages this insight for one-shot visual imitation learning of bimanual manipulation from a human demonstration.
Our method uses a single demonstration as input as an RGB-D video of a human performing a bimanual manipulation task. \methodname{} interprets the demonstration as a screw motion between both hands and uses the perceived bimanual grasp and virtual joint to train a prediction model on 3D point clouds.
This model predicts full bimanual manipulation behaviors composed of bimanual grasping strategies and two-arm relative motion in a possibly moving reference frame, for novel views of the object.
These predictions form the starting hypothesis for a self-practicing iterative process. Here, the robot engages in bimanual interactions, learning to overcome morphological differences by optimizing a reward signal generated autonomously, resulting in successful bimanual manipulation strategies.
The new strategy can then be used in a self-improving loop to retrain a better prediction model that is also able to generalize to new instances of the same object class thanks to a set of geometric augmentations.

We demonstrate the performance of our solution in six challenging bimanual manipulation tasks involving different types of screw motion between both hands, both in objects with physical kinematic constraints and in tasks where the constraints need to be virtually created by the agent. 
Our experiments indicate that the projection into the screw-axis space is a robust representation for bimanual manipulation---leading to sample efficient exploration and strong performance in executing bimanual tasks.
\section{Related Work}
\label{s_rw}

\methodname{} is a novel solution to generate and refine autonomous robot bimanual manipulation behavior bootstrapped with a single video of a human demonstration. In the following, we contrast \methodname{} to the most relevant prior work in robot bimanual manipulation and visual imitation learning.

\paragraph{Bimanual Manipulation}
Early on, robotics researchers acknowledged the need for bimanual manipulators to solve tasks in unstructured environments~\cite{goertz1952fundamentals,ambrose2000robonaut,5980058}.
Generating coordinated behavior for both arms became a significant challenge~\cite{smith2012dual, billard2019trends} that researchers have attempted to solve with planning~\cite{koga1992experiments,vahrenkamp2009humanoid, cohen2014single}, control~\cite{hsu1993coordinated}, reinforcement learning~\cite{colome2020reinforcement}, and imitation learning~\cite{zollner2004programming, kim2021transformer}.
To generate bimanual manipulation behavior, these solutions have to explore a large action space with strict temporal and spatial synchronization. 
A common strategy is to coordinate behavior using stable static postures or keypoints~\cite{gribovskaya2008combining, asfour2008imitation, silverio2015learning}, or with explicit spatial or temporal constraints typically extracted via kinesthetic teaching~\cite{calinon2012statistical, gams2014coupling, ureche2018constraints, figueroa2017learning}. 
Oftentimes, these approaches necessitate specialized teleoperation hardware like custom devices~\cite{pais2015learning, zhao2023learning, tung2021learning} or motion capture~\cite{krebs2022bimanual, dreher2022learning} that limit their scalability and availability. 
In contrast, \methodname{} uses a single RGB-D video of a human demonstration, which is cheaper to acquire, scalable and does not require controlling a robot.

Given the large action space and difficulty of exploration, reinforcement learning approaches to bimanual manipulation are prohibitively costly to train on real robot hardware. As an alternative, researchers have explored sim-to-real approaches~\cite{chen2022towards, kataoka2022bi, chitnis2020efficient}. These approaches suffer from the reality gap which is exacerbated in contact-rich manipulation tasks with complicated dynamics~\cite{zhao2020sim}. 
An alternative approach is to employ movement primitives~\cite{lioutikov2016learning, avigal2022speedfolding,  xie2020deep, weng2022fabricflownet, ha2022flingbot, chitnis2020intrinsic, grannen2020untangling, grannen2022learning, batinica2017compliant, franzese2023interactive, ganapathi2021learning, amadio2019exploiting, bahety2023bag} which reduce the search space but limit expressiveness and typically require substantial engineering effort. 
Recently, \citet{grannen2023stabilize} proposed a stabilizing-acting bimanual manipulation framework where the stabilizing hand is trained using human annotations and the acting hand is trained using kinesthetic demonstrations. In contrast, \methodname{} leverages a novel action representation that efficiently learns bimanual manipulation policies given only a \textit{single human video} demonstration, and can correct failed actions through a self-supervised policy fine-tuning method.

\paragraph{Visual Imitation Learning}
Recent work has sought to imbue robots with the ability to learn from large collections of unstructured human videos like Ego4D~\cite{grauman2022ego4d} or YouTube videos \cite{sivakumar2022robotic, xiong2021learning}. 
Some works have proposed learning cost functions from video and language data \cite{chen2021learning, shao2021concept2robot, ma2023liv, ma2022vip}, whereas others propose pretraining objectives~\cite{nair2022r3m, radosavovic2023real}. More direct approaches generally track human hands in videos (e.g. with FrankMocap~\cite{rong2021frankmocap}), mapping the hand trajectories to the robot's action space \cite{mandikal2022dexvip, shaw2023videodex, shaw2024learning, kannan2023deft}. A common approach in these works is to structure video understanding by modelling manipulation using affordances (i.e. detecting contact points \cite{shan2020}) and subsequent interaction trajectories .
Since the robot's embodiment differs from a human demonstrator and tracking is generally noisy, interactive fine-tuning is generally necessary to obtain a reliable behavior policy~\cite{bharadhwaj2023zeroshot, bahl2022human, mendonca2023structured}. 
DEFT~\cite{kannan2023deft}, for example, trains an affordance prediction model on large-scale data and obtains the interaction trajectory given a human demonstration at test time. This trajectory is then refined through interaction. 
Inspired by this line of work, we propose a novel formulation of synchronized bimanual manipulation learned solely by watching human-object interactions in video. In contrast to prior work in unimanual manipulation, our focus here is on the action representation. Our unique formulation of bimanual motion in terms of screw joints abstracts complex high-DoF manipulation into a unified framework---enabling efficient imitation learning from video.

\section{Preliminaries: Screw Theory}
\label{s_prelim}

\begin{figure*}[t]
    \centering
    \includegraphics[width=1\textwidth]{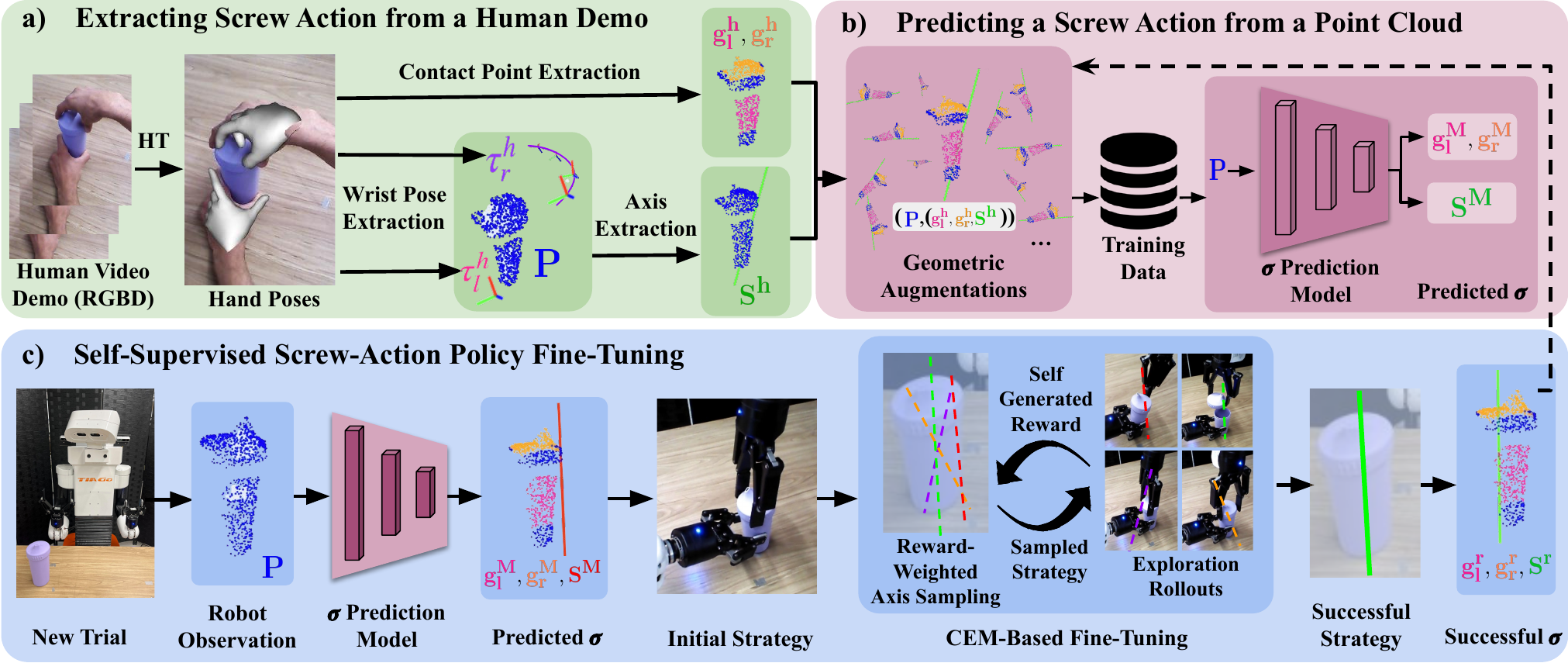}
    \caption{\textbf{Overview of \methodname{}.} \textbf{a)} Given an RGB-D video of a human performing a bimanual task, we use off-the-shelf hand tracking (HT) models~\cite{rong2021frankmocap,shan2020} to extract a trajectory of wrist poses $\tau^h$ and grasp contact points $(g_l^h,g_r^h)$. \methodname{} interprets $\tau^h$ as a screw motion between both hands to estimate screw axis parameters $S^h$ (Sec.~\ref{sec:method_A}). \textbf{b)} Next, we apply geometric augmentations on the 3D object point cloud to train a PointNet~\cite{qi2016pointnet} model to estimate screw actions for novel object views (Sec.~\ref{s_prediction}). \textbf{c)} Finally, the trained model generates an initial hypothesis that the robot executes and iteratively refines using an autonomously generated reward signal. The successful data point is further used to improve the prediction model (Sec.~\ref{sec:fine-tuning}).}
    \label{fig:method_diagram} 
\end{figure*}

\methodname{} models bimanual manipulation as a screw motion between the two hands. Chasles' theorem states that any rigid body motion can be written as the composition of a rotation of the body about a unique line in space and a translation along the same line. This line is referred to as the \textit{screw axis} of that motion. A screw axis $\mathcal{S}$ can be represented as $(q, \hat{s}, h)$ where $q \in \mathbb{R}^3$ is any point on the axis, $\hat{s} \in \mathbb{R}^3 $ is a unit vector in the direction of the axis, and $h \in \mathbb{R}_+$ is the pitch of the screw, defining the ratio of linear motion along the screw axis to the rotational motion around the screw axis~\cite{siciliano2008springer,lynch2017modern}. 

Assuming some angular displacement $\theta \in \mathbb{R}$ along a screw axis $\mathcal{S}$, the corresponding rigid body motion in exponential coordinates, $\xi = (\omega, v)\in \mathbb{R}^6$ is given by 
\begin{align}
\label{eq:screw}
    \xi= \mathcal{S}\theta = & \begin{bmatrix}
        \omega \\
        v 
    \end{bmatrix} = 
    \begin{bmatrix}
      \hat{s} \theta \\
       - \hat{s} \theta \times q  + h \hat s \theta
    \end{bmatrix}
\end{align} 
where $\omega \in \mathbb{R}^3$ represents angular motion, 
and $v \in \mathbb{R}^3$ represents linear motion.
To transform this into a homogeneous transformation matrix, $T\in SE(3)$ ($SE(3)$ the Special Euclidean Lie Group) we apply the matrix exponential: $T=\exp([\mathcal{S}]\theta)$, where $[\mathcal{S}]\theta \in se(3)$ ($se(3)$ is the Lie algebra) is the matrix representation of the exponential coordinates:
\begin{align}
\label{eq:angular_vel}
    & [\mathcal{S}]\theta = \begin{bmatrix}
        [\omega] & v\\
        0 & 0
    \end{bmatrix}. 
\end{align}
with $[\omega] \in so(3)$ is a skew-symmetric matrix representing orientation.

Conversely, given a rigid body transformation, $T\in SE(3)$, we can compute the corresponding screw axis as follows. In the case of a pure translation ($h=\infty$), the screw axis can be recovered as $\hat{s}$ pointing in the direction of linear motion, and $q$ is any point. In the case of pure rotation ($h=0$), we can recover the corresponding twist in matrix form $[S]$ using the matrix logarithm: $[\mathcal{S}]\theta = \log(T)$.
Applying Eq.~\ref{eq:screw}, we can obtain the screw axis parameters as
\begin{align}
\label{eq:screw_sq}
    & \hat{s} = \frac{\omega}{||\omega||}, 
    \quad
    q = \frac{\hat{s} \times v}{||\omega||}.
\end{align}
For the general case with $h \notin \{0, \infty \}$ and further details, we refer the reader to \citet{lynch2017modern}. 


We will use the definitions above to infer a screw axis from a sequence of relative transformations between human hands, and to generate relative motion between robot hands for a given screw axis.
In this work, we will consider three screw types: pure translation ($h=\infty$, \texttt{prismatic}), pure rotation ($h=0$, \texttt{revolute}), and rotation with a fixed orientation (\texttt{revolute3D}). We describe the axis computation and trajectory generation for each case in Sec.~\ref{s_system} below. 

\section{\methodname{}: Policy Learning with Screw Actions}
\label{s_system}

Seminal research in psychology and biomechanics~\cite{guiard1987asymmetric} indicates that bimanual behavior in humans can be modeled as if ``a serial kinematic chain would connect both hands'', where one hand (left) sets a spatial reference frame and the other (right) moves relative to it. 
Inspired by this work, we propose a novel action space parametrization for robotic bimanual manipulation that we call \textbf{screw actions}, that fully specifies the behavior of both hands through a screw joint  between the hands. A screw action is defined as, $\sigma = (g_l, g_r, S, \tau_l)$ in its most general form. $g_l$ and $g_r$ are the grasping/placing locations for left and right hands. $S$ is a 1-DoF screw axis describing the relative motion between left and right hands. Finally, $\tau_l$ is a possible sequence of left-hand pose changes during the interaction (e.g. moving a pot to the stove while stirring) that can be empty if the left hand just fixates/stabilizes the object.

Given a screw action, the motion of both hands of the robot during the bimanual manipulation is fully specified. Our main hypothesis is that our new action space simplifies robot dexterous bimanual manipulation at two levels: first, it aids in learning from visual human demonstrations by projecting noisy multi-hand motion into a simpler constrained space, and second, it facilitates fine-tuning the perceived motions by providing a constrained space in which real-world exploration is more efficient. We propose a novel solution, \methodname{}, that leverages this insight to learn bimanual policies. \methodname{} integrates three modules: a perceptual module to interpret a single human demonstration as a screw action, a prediction model that predicts screw actions based on a point cloud of an object, and a self-supervised iterative fine-tuning algorithm that explores in screw action space to find optimal parameters for bimanual tasks. In the following, we explain each of these modules in detail.

\subsection{Extracting a Screw Action from a Human Demonstration}
\label{sec:method_A}
The first module of \methodname{} (Fig.~\ref{fig:method_diagram}a) parses an RGB-D video of a human demonstrating a bimanual task into a suitable action representation for robot execution, in our case, a screw action $\sigma^{h} = (g_l^{h}, g_r^{h}, S^{h}, \tau_l^{h})$ ($h$ indicates \textit{human}). \methodname{} first extracts the grasping/placing location of the human hands ($g_l^h$ and $g_r^h$) using an off-the-shelf hand-object detector~\cite{shan2020}, detecting the first intersection of the hand and the object bounding boxes in the RGB image sequence, and projecting it into the 3D point cloud of the object, $P$, using the information of the depth channel.

\methodname{} then extracts the 6-DoF trajectories (position and orientation) of the human hands using an off-the-shelf hand-tracking solution (FrankMocap~\cite{rong2021frankmocap}) to detect the wrist poses over time, $\tau_l^{h}$ and $\tau_r^{h}$.
A direct approach would use these trajectories to imitate and fine-tune the bimanual manipulation. 
However, the original trajectories contain noise from the visual tracker, the motion of both hands is not constrained to be synchronized, and an embodiment gap exists between the human hand and the robot gripper, making it harder to imitate and fine-tune (as shown in Sec.~\ref{s_exp}); \methodname{} overcomes these limitations by interpreting the trajectories as a screw action.

\begin{figure}[t]
    \centering
    \includegraphics[width= 0.97\linewidth]{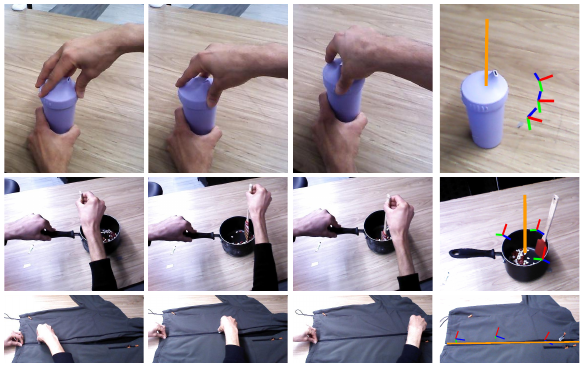}
    \caption{\textbf{Human demonstrations as screw actions.} Three frames of a human demonstration for three bimanual tasks (top row: opening a bottle, $m=$~\texttt{revolute}, middle row: stirring a pot, $m=$~\texttt{revolute3D}, bottom row: opening a zipper, $m=$~\texttt{prismatic}) and the perceived screw axis explaining the motion (fourth column, orange indicates the axis line). Our screw action representation facilitates the interpretation of noisy hand trajectory observations in a bimanual interaction as evidence of a simple 1-DoF constraint between both hands}  
    \label{fig:screw_demos}
    \vspace{-1em}
\end{figure}

Inspired by models of human bimanual manipulation~\cite{guiard1987asymmetric}, we assign an acting and reference role to the right and left hands, respectively, keeping $\tau_l^{h}$ as the trajectory of the left hand.
\methodname{} finds then the screw axis $S^h$ by transforming $\tau_r^{h}$ to the left hand reference frame and analyzing the left-right relative motion to obtain the screw-joint type, $m$, and parameters, $\hat{s}$ and $q$.
For that, \methodname{} assumes three possible screw joint types, prismatic, revolute and revolute with fixed orientation, as explained in Sec.~\ref{s_prelim}. Assuming $m=$~\texttt{prismatic}, \methodname{} obtains the screw parameters by fitting a 3D line to the trajectory of the right wrist. When $m=$~\texttt{revolute}, the screw axis parameters can be obtained by transforming each pose of the right wrist relative to the left wrist to exponential coordinates using the matrix logarithm, applying Eq.~\ref{eq:screw_sq}, and averaging the resulting $\hat{s}$ and $q$. Finally, if $m=$~\texttt{revolute3D}, \methodname{} first fits a plane to the trajectory of the right wrist; the normal to the plane provides, $\hat{s}$. The right wrist trajectory is then projected onto the plane and \methodname{} fits a circle to it; the center of the circle provides $q$. 
We employ a Maximum a Posteriori Estimation (MAP) method to determine the screw joint type ($m$) corresponding to a human demonstration. In this case, we want to estimate $m$ based on the hand pose observations, $\tau^h_r$. To do this, \methodname{} evaluates the likelihood of each joint type ($m$), by comparing the observed demonstration trajectory $\tau^h_r$ with the trajectory computed based on $m$. This comparison involves evaluating a score function that measures the distance between the two trajectories, considering both positional and angular differences at each waypoint. A lower distance indicates greater similarity between trajectories. \methodname{} picks the joint type with the highest likelihood.
Examples of the extracted screw axes for each type are depicted in Figure~\ref{fig:screw_demos}. 

\subsection{Predicting a Screw Action from a Point Cloud}
\label{s_prediction}
Once the robot perceives the human demonstration of the bimanual task and represents it in the screw action space, how can it generalize to novel object instances and configurations? 
To tackle this, the second module of \methodname{} (Fig.~\ref{fig:method_diagram}b) includes a PointNet~\cite{qi2016pointnet} based model trained to predict the screw action from an object's point cloud.
Concretely, given an RGB-D observation, we use MDETR~\cite{kamath2021mdetr} to segment out the object and extract its partial point cloud, $P$. The goal is to learn a perception model $M:P \mapsto \sigma^{M}=(g_l^{M}, g_r^{M}, S^{M})$ ($M$ indicates \textit{Model}).
Here $g_l^{M}, g_r^{M}$ refer to the grasp contact points predicted by the perception model for the left and the right grippers respectively and  $S^{M}$ refers to the predicted screw axis.
From here on, we omit the left-hand trajectory, $\tau_l$, since our experiments focus on learning the relative motion between hands, but the left-hand motion is enabled by our general formalism, as shown in additional trials in Appendix \ref{sec:left_tau} and website.


\methodname{} benefits from the 3D nature of both the input observation and screw action representation that allows for straightforward geometric augmentations of the data: translation, rotation, and scaling. These augmentations are applied to the point cloud, $P$, and corresponding robot action $\sigma^{M}$. As a result, we generate an extensive training dataset from just a \textit{single} human demonstration.
Using PointNet~\cite{qi2016pointnet} as the backbone, we construct two networks: a regression network trained with MSE loss to predict the axis and a segmentation network trained with negative log likelihood loss for identifying contact points.
We train task-specific prediction models. 
The training of each model is efficient, requiring on average 40 minutes for 2000 epochs on a RTX 4090 GPU. This rapid training cycle enables quick model refinement and incorporation of new data, as we will discuss in the next section.

\subsection{Self-Supervised Screw-Action Policy Fine-Tuning}
\label{sec:fine-tuning}
Given that human pose tracking is inherently noisy, the prediction model trained on the human hand trajectory will necessarily have some error.  Thus, if the robot directly executes the bimanual manipulation defined by the predicted screw action, it will probably fail (as evidenced in our experiments). Nevertheless, the predicted screw action produces a behavior close to a successful manipulation and thus can be used as initialization for a fine-tuning procedure through interaction. The third module of \methodname{} consists of a self-supervised policy improvement algorithm that refines the noisy screw action (Fig.~\ref{fig:method_diagram}c). As our experiments indicate (Sec.~\ref{s_exp}), the use of screw actions as policy parameterization is critical for more efficient bimanual exploration and allows \methodname{} to achieve success in multiple tasks. In the following, we first explain how a screw action is used by \methodname{} to generate a bimanual manipulation behavior, and then, we describe the iterative process to fine-tune an initial (failing) screw action into a successful one.

Given a predicted screw action $\sigma^M = (g_l^M, g_r^M, S^M)$, 
the two grippers first go to $g_l^M$ and $g_r^M$ at pre-defined orientations using a whole-body controller ~\cite{mansard2009versatile}.
The end of this initial motion is the beginning of the bimanual manipulation described in Section \ref{s_prelim} by the screw axis $S$. While the left hand is possibly executing a trajectory $\tau_l$, the right hand will move relative to it following the constraints indicated by $S$. \methodname{} creates $k\in{1\ldots K}$ waypoints along the screw axis with steps of $\theta_T/K$, where $\theta_T$ is a pre-determined total amount of translation along the axis-line for $m=$~\texttt{prismatic} type, or the total amount of rotation around the axis-line for $m=$~\texttt{revolute} and $m=$~\texttt{revolute3D} types. In the latter case, the orientation of the right hand is kept constant during the motion. Assuming an initial 6D pose for the right hand of $T_0^{right}$ with respect to the left hand, the right-hand poses will be given by $T_i =\exp([S]\theta_k)\ T_{0}^{right}$, where $\exp$ is the matrix exponential and $\theta_k = k\theta_T/K$.

  

Given the method explained above to generate bimanual manipulation behavior based on a screw action, we now explain the iterative procedure to fine-tune an initially failing action. Inspired by prior exploratory approaches~\cite{kannan2023deft, bahl2022human, stulp2012path}, \methodname{} implements a sampling-based optimization framework based on the cross-entropy method (CEM).
The process starts with obtaining the initial screw action from the prediction model that was trained on the human demonstration. Next, an initial sampling distribution, $D$, is used to sample screw axis parameters around the initial screw axis.
$E$ samples, $\xi_{1, e}$, are drawn from this distribution. 
CEM then requires a reward to score each sample and guide a reweighting of the sampling distribution ($D$) for the next epoch.
In \methodname{}, the CEM optimization process is self-supervised through an autonomously generated reward based on the length of the episode and the amount of force employed, measured by a force-torque (FT) sensor in the right hand's wrist. 
Concretely, after each epoch, 
all the trajectories up to that epoch are ranked by their length: the longer an episode runs without failure, the better it is. 
We implement three self-detected failures: 1) when the robot is not applying enough force (norm of the wrench signal is below a threshold), indicating that it may be moving in free space instead of manipulating, 2) when the robot is applying too much force (norm of the wrench signal is above a threshold), indicating that it is trying to manipulate an object in the wrong way, and 3) when the robot loses grasp (measured by the finger proprioception).
After all the episodes are ranked by their lengths, \methodname{} takes the top $T$ trajectories and ranks them by the mean wrenches employed over the episode; using lower force for the manipulation is considered more efficient. These episodes form the elite set. \methodname{} updates the sampling distribution based on the elite set and uses the new distribution in another epoch. The process repeats for $N$ epochs or until the bimanual manipulation succeeds. The CEM fine-tuning procedure is summarized in Algorithm~\ref{alg:1}.

The successfully executed screw action $\sigma^r = (g_l^r, g_r^r, S^r)$ is added to the training dataset (see Fig.~\ref{fig:method_diagram}) to enhance the action prediction model. This iterative process, if performed repeatedly, can facilitate continuous improvement of both the robot's policy and the prediction model, creating a self-supervised feedback loop where each component bolsters the other as demonstrated in Sec.~\ref{s_exp}.

\begin{algorithm}
\caption{Cross-Entropy Method Optimization}\label{alg:cap}
\begin{algorithmic}
\Require parameter distribution $D$, total epochs $N$, episodes in each epoch $E$, elite trajectories threshold $T$, $S_{init} = (\hat{s}_{init}, q_{init})$ initial screw axis\\
\State $\xi_{init} \gets (\hat{s}_{init}, q_{init})$
\State $D \gets N(0, \sigma^2)$
\For{$n = 1 ... N $}
\For{$e = 1 ... E $}
\State Sample $\epsilon_{n, e} \sim D$
\State Execute $\xi_{n, e} = \xi_{init} + \epsilon_{n, e}$
\State Collect reward $R_{n, e}$; reset environment
\EndFor
\State ${\xi_1, \xi_2 ... \xi_T} \gets$ Order trajectories $\xi_{0,0}, \xi_{0,1}$ ... $\xi_{n, E}$ based on rewards
\State $\Omega \gets \{\epsilon_{\xi_1}, \epsilon_{\xi_2}$ ... $\epsilon_{\xi_T} \}$ 
\State Fit $D$ to $\Omega$ 
\EndFor
\State $\xi_{final} \gets \xi_{init} + \epsilon_{final}$
\end{algorithmic}
\label{alg:1}
\end{algorithm}

\section{Experimental Evaluation}
\label{s_exp}

\begin{figure*}[t]
    \centering
\includegraphics[width=1\textwidth]{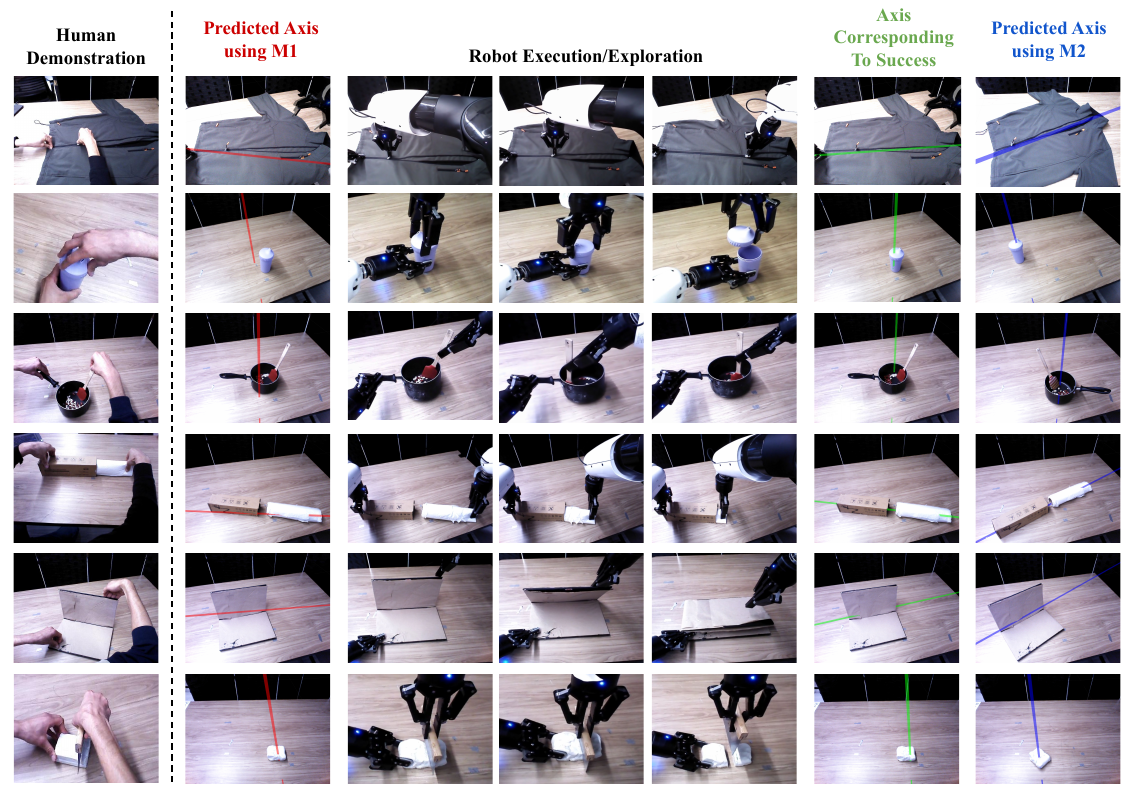}
    \caption{\textbf{Screw Action Fine-tuning and Prediction Model Re-training Result.} The first column shows the human demonstration for each task. The second column shows the axis predicted by M1, the model trained on the axis extracted from the human demonstration, with the object at a novel pose. Columns 3-5 show snapshots from an episode in the fine-tuning stage. Column 6 shows the axis corresponding to the successful trajectory obtained during the aforementioned process. Column 7 shows the predicted axis for a novel object pose from the prediction model re-trained on the corrected axis. This result shows how the robot starts from a noisy screw axis and using the screw action fine-tuning, corrects the axis. Furthermore, it also shows that this corrected axis can be used to re-train the prediction model to output a more accurate axis.}
    \label{fig:results_main} 
    \vspace*{-3mm}
\end{figure*}

\begin{figure*}[t]
    \centering
    \includegraphics[width=1\textwidth]{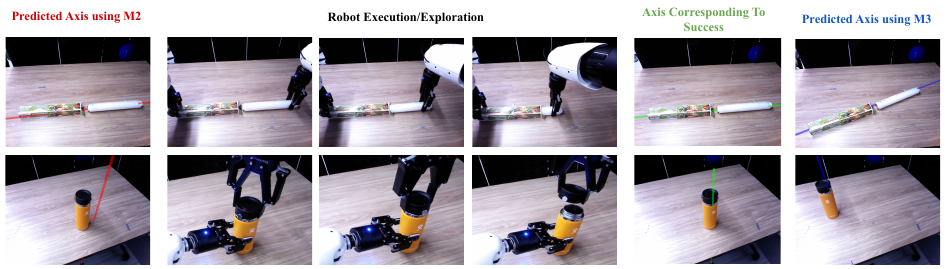}
    \caption{\textbf{Generalization to New Objects.} The first column shows the axis predicted by M2, the model trained on the corrected screw action for the first object. Columns 2-4 show snapshots from an episode in the fine-tuning stage. Column 5 shows the axis corresponding to the successful trajectory obtained during the aforementioned process. Column 6 shows the predicted axis from the prediction model re-trained on the corrected axis (M3). Thus, \methodname{} can obtain reasonable screw action predictions and fine-tune them to generalize to new objects.}
    \label{fig:results_gen} 
    \vspace*{-1em}
\end{figure*}

We evaluate \methodname{} on six real-world bimanual tasks: \texttt{open bottle}, \texttt{close zipper}, \texttt{insert roll}, \texttt{close laptop}, \texttt{stir} and \texttt{cut}. 
These tasks collectively encompass three types of screw joint models: \texttt{prismatic}, \texttt{revolute}, and \texttt{revolute3D}. They also involve screw actions in two types of objects: articulated objects with actual physical joints constraining their motion (as in bottles, rolls, and laptops) and objects without constraints, where the screw action creates a virtual joint that facilitates the correct bimanual manipulation (as in stirring, cutting, and zipping a jacket). While the manipulation of articulated objects has been studied more extensively in the past~\cite{sturm2010operating,karayiannidis2016adaptive,martin2017cross,abbatematteo2019learning,wang2022adaafford}, this is the first time, to the best of our knowledge, that a framework unifies the bimanual manipulation of rigid and articulated objects through virtual joints.
In the following, we explain each task in brief:

\begin{itemize}[leftmargin=*]
    \item \texttt{open bottle}: A bottle with its cap closed is placed upright on the table. The robot performs the opening action as defined by the screw action, followed by a lift arm command. We consider success if the cap is separated from the base of the bottle at the end.
    \item \texttt{close zipper}: A jacket is kept in a configuration as shown in the first row of Fig. \ref{fig:results_main}. We consider success if the robot zips 90\% of the jacket at the end.
    \item \texttt{insert roll}. A roll is placed beside the box aligned as shown in the fourth row of Fig. \ref{fig:results_main}. We consider success if the robot inserts 90\% of the roll inside the box at the end. 
    \item \texttt{close laptop}: A laptop is placed on the table, opened to around $100\degree$. We consider success if the robot closes the laptop (final opening $< 10^\circ$).
    \item \texttt{stir}: A container with a ladle propped against its side is placed in front of the robot. The container has two different colored beans, initially separated. We consider success if the two types of beans are significantly mixed after the stirring as measured by a human evaluator.
    \item \texttt{cut}: The robot is holding a scraper knife in one gripper and tasked with cutting a block of clay ($\sim$7 cm in height). We consider success if the block of clay is cut into two pieces at the end.
\end{itemize}

In all our experiments, we use a PAL-Robotics Tiago++ bimanual manipulator and control its two arms using a whole-body controller that maps desired end-effector poses for both arms to joint torques using an inverse-kinematics-based solution with task-priority control to avoid self-collisions~\cite{mansard2009versatile}. For perception, we use an Orbbec Astra S RGB-D camera mounted on Tiago++'s head both to observe humans and to predict screw actions on objects, and an ATI mini45 force-torque sensor mounted on the right hand's wrist.

For each task, \methodname{} begins with the screw action predicted by the trained model after observing a single human interaction using a perceived point cloud as input, and fine-tunes it using its self-supervised iterative procedure. 
Each trial of the procedure is limited to a maximum of 5 epochs, each containing 5 episodes, after which, if the procedure did not find a successful screw action, we consider the trial a failure.
The fine-tuning takes around 40 minutes, demonstrating a reasonable real-world exploration time.
Success is verified manually after each episode, and a human resets the environment if necessary.

\noindent
\subsection*{Experiments and Results:}

In our experiments, we aim to answer four questions:

\vspace{0.5mm}
\textit{Q1) Is a single human demonstration enough for \methodname{} to achieve success in bimanual manipulation tasks?} 
To evaluate this question, we perform three trials per human demonstration for each of the tasks and observe if, in the trials, \methodname{} successfully achieves the bimanual tasks with its self-supervised fine-tuning in screw action space. We also annotate the amount of interaction (episodes) necessary on average to succeed in the task. We use a single demonstration per task, but each trial starts with a different (novel) location of the object(s). Therefore, \methodname{} needs to predict the screw action in a new location and start the iterative process there. A trial for each task is depicted in each row of Fig.~\ref{fig:results_main}, columns 1 to 6. The results are summarized in Table~\ref{Tab:main}.

\begin{table}[h]
    \begin{center}
        \caption{Generalization to new object poses}
        \begin{tabular}{ l|cc }
            \toprule
            
              & \# Successes & Avg Epochs and Episodes         \\
            \midrule
             Open Bottle & 2/3 & (3, 18)\\
             Close Zipper & 3/3 & (2, 11) \\
             Insert Roll & 3/3 & (1, 8) \\
             Close Laptop & 3/3 & (2, 12) \\
             Stir & 2/3 &   (3, 16)\\
             Cut & 3/3 &  (1, 7)\\
            \bottomrule
        \end{tabular}
        \label{Tab:main}
    \end{center}
    \vspace{-1em}
\end{table}


Overall, \methodname{} achieves an aggregated success of ~90\% in all trials. \methodname{} failed only in one trial of the \texttt{open bottle} and the \texttt{stir} tasks as the fine-tuning process finished without any successful screw action. Due to the small number of episodes (25 maximum), we observe a marked dependency on the first screw action samples for the fine-tuning procedure, which could be alleviated with a larger number of episodes per epoch. Despite that, we consider that our experiments indicate that, in most cases, \textbf{\methodname{} succeeds in all studied bimanual manipulation tasks using only a single video of a human demonstration}, thanks to the structure provided by the screw action for perception and self-supervised exploration. 
Additionally, we also conduct experiments to analyze the robustness of \methodname{} to noisy demonstrations. We observe that despite noisy demonstrations, \methodname{} is able to extract a screw axis sufficiently accurate for fine-tuning. The details and results of the experiment are shown in Appendix. \ref{sec:noisy_demo}


\vspace{2mm}
\textit{Q2) Can a policy fine-tuning and model retraining loop enable \methodname{} to continually improve and generalize to new objects?}
We assess if \methodname{} can use the corrected screw action obtained after fine-tuning to improve the prediction model and generalize to unseen objects. In this experiment, the screw action prediction model is first trained with the noisy screw action parsed from the human demonstration (denoted M1 in Table~\ref{Tab:gen}). The robot then executes and fine-tunes this screw action to obtain a corrected one and uses it to re-train the prediction model, obtaining the model M2. \methodname{} then uses model M2 to predict and execute the bimanual tasks and performance is measured. Table~\ref{Tab:gen} reports the exploration iterations required until success is achieved as (epochs, episodes), where each epoch consists of 5 episodes and the policy is updated after each epoch. Our experiments indicate that, after retraining, \methodname{} succeeds at the task with the same object (second column) almost zero-shot, showing that the prediction model can be iteratively improved using the corrected action obtained from the fine-tuning stage. 

We then assess how \methodname{} handles new objects. We place a novel object of the same category at a new pose and run the same experiment (see Fig.~\ref{fig:results_gen}). For the \texttt{open bottle} task, initially, using M2, the screw action prediction is sub-par (Table~\ref{Tab:gen}, third column). This is due to the large geometric difference between the bottle that M2 was trained on and the new bottle. However, after fine-tuning and obtaining model M3, \methodname{} can complete the task with the new object almost zero-shot (Table \ref{Tab:gen}, fourth column). For the \texttt{insert roll} task, the initial screw action prediction is good as the structural difference is smaller between the old and the new object. This indicates that \textbf{\methodname{} helps create a self-learning loop where the robot can continually expand its manipulation capabilities to new objects}. We also compare training from scratch with a pre-trained PointNet model and observe an improvement in the screw
axis prediction on novel objects. The details and results are shown in Appendix. \ref{sec:pretrained_pointnet}

\begin{table}[h]
    \begin{center}
        \caption{Generalization to new objects $^1$}
        \begin{tabular}{ l|cc|cc }
            \toprule
             & \multicolumn{2}{c}{Same object} & \multicolumn{2}{c}{New object} \\
             & M1 & M2 & M2 & M3       \\
            \midrule
             Open Bottle & (3, 16)  & (0, 1)& (2, 14) & (0, 1)\\
             Insert Roll & (1, 7 )& (0, 2)& (0, 3 )& (0, 1)\\
            \bottomrule
        \end{tabular}
        \label{Tab:gen}
    \end{center}
    \footnotesize{$^1$ Results reported as (Epochs, Episodes) until a success is reached. }
    \vspace{-3mm}
\end{table}
\begin{table}[t!]
\caption{Action Representation Comparison} \vspace{-3mm}
\centering
\begin{tabular}{ ll|ccc} 
    \toprule
        & \multirow{2}{*}{Task} & \multicolumn{3}{c}{} \\ 
        & & Success? & \#Episodes & Dense Metric \\
    \midrule
    FM + $N\times$6-DoF & Bottle & No & 25/25 & 0\%\\
    (DEFT*)  & Roll & No & 25/25 & 0\% \\
        & Laptop & No & 25/25 & 10\% \\
    \midrule
    Screw + $N\times$6-DoF & Bottle & No & 25/25 & 10\%\\
        & Roll & No & 25/25 & 50\%\\
        & Laptop & No & 25/25 & 50\%\\
    \midrule
    Screw + Screw & Bottle & Yes & 16/25 & 100\%\\
    (\methodname{}) & Roll & Yes & 7/25 & 100\%\\
        & Laptop & Yes & 11/25 & 100\%\\
    \bottomrule  
\end{tabular}
\label{Tab:ablation_repr}
\vspace{-3mm}
\end{table}

\begin{figure}[t!]
    \centering
    \includegraphics[width=\linewidth]{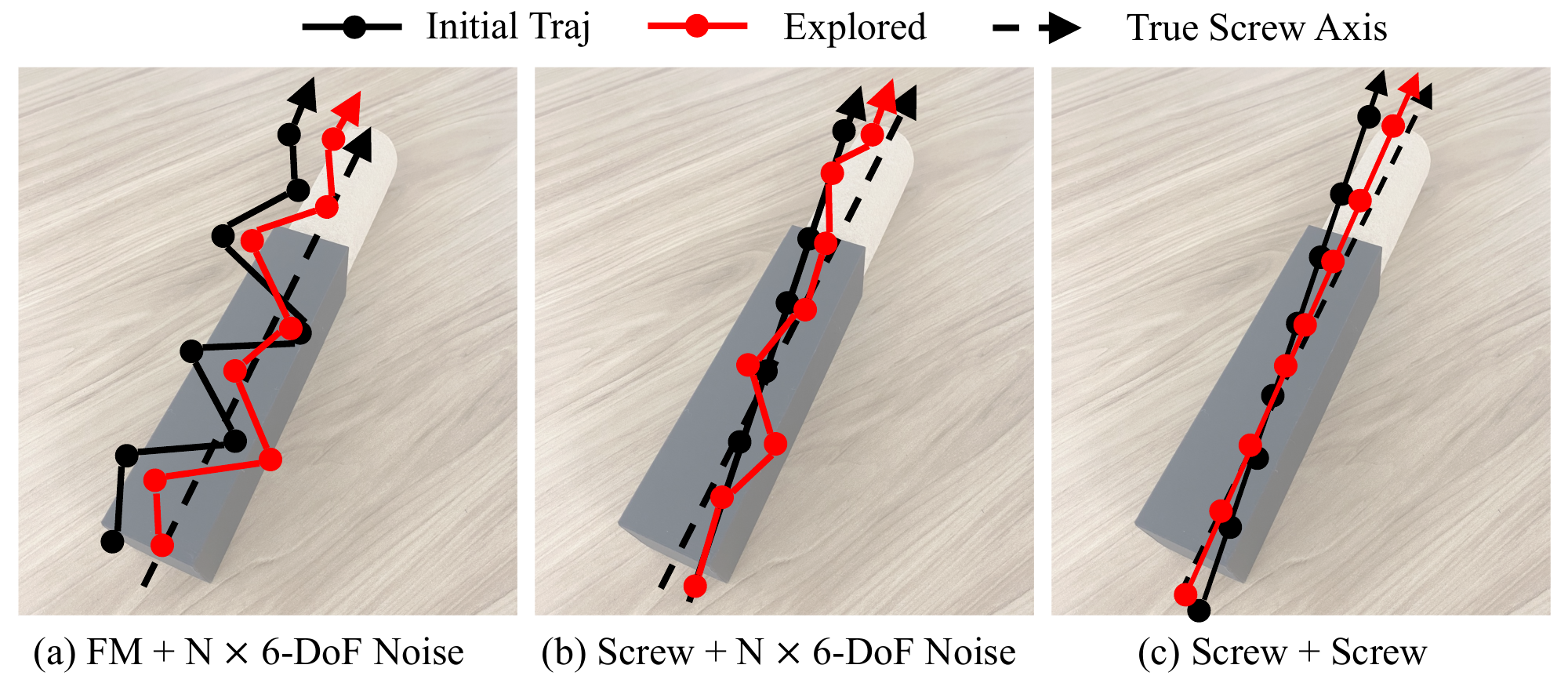}
    \vspace{-8mm}
    \caption{\textbf{Exploration Intuition.} The different action representations from Table~\ref{Tab:ablation_repr} are illustrated in 2D for the \texttt{insert tissue roll} task. The true (prismatic) screw axis is visualized as a dashed line. The resulting initial and sampled exploration trajectories are visualized as black and red, respectively. (a) The baseline \textit{FM + $N\times$6-DoF}: obtaining an initial trajectory from FrankMocap as $N$ 6-DoF waypoints, then performing exploration around that trajectory with noise in 6-DoF space. (b) The baseline \textit{Screw + $N\times$6-DoF}: perceiving the initial trajectory as a screw axis but exploring with noise in 6-DoF space. (c) \methodname{} (\textit{Screw + Screw}), perceiving the demonstration as a screw action and exploring in the space of screw axes. }  
    \label{fig:screw_expl}
\end{figure}

\vspace{2mm}
\textit{Q3) What benefits does the screw axis representation have compared to a more direct, $N\times$6-DoF representation?}
To assess the importance of the screw axis representation we compare it with two baselines visualized in Fig.~\ref{fig:screw_expl}. First, the \textit{FM + $N\times$6-DoF} baseline (Table \ref{Tab:ablation_repr}, first row) extracts the initial trajectory from hand tracker using the wrist poses as $N$ waypoints directly. We call this space as $N\times$6-DoF as it has $N$ waypoints with each waypoint described by a 6-DoF pose. 
During fine-tuning, it explores in the $N\times$6-DoF space by adding noise to the initial waypoints (Fig.~\ref{fig:screw_expl}a). With this baseline, we ablate the screw representation both as the human demonstration parser and as the action space during fine-tuning. 
Note that this baseline emulates DEFT’s \cite{kannan2023deft} fine-tuning stage. Due to the unavailability of DEFT's code/model, we attempt to approximate \textsc{DEFT}’s methodology as closely as possible. Key differences include: 1) the use of both hands in our method, 2) the extraction (rather than prediction) of grasping locations directly from demonstrations, and 3) a reliance on \textsc{ScrewMimic}’s self-generated CEM reward, rather than human-assigned scores.
The second baseline, \textit{Screw + $N\times$6-DoF} (Table \ref{Tab:ablation_repr}, second row) extracts the initial trajectory from the output of the hand tracker using \methodname{}'s parser module as a screw axis, but explores by adding Gaussian noise in SE(3) during fine-tuning (Fig.~\ref{fig:screw_expl}b). This baseline parses the human demonstration in the same way as \methodname{} but explores differently.
We compare against our proposed \methodname{} (Table \ref{Tab:ablation_repr}, third row), indicated as \textit{screw + screw} (Fig.~\ref{fig:screw_expl}c). 

Each baseline obtains an initial trajectory from a human demonstration, then performs the fine-tuning procedure in the corresponding action space. 
For each method, we run one trial of the exploration process for each of the three tasks ---\texttt{open bottle}, \texttt{insert roll} and \texttt{close laptop} as shown in Table \ref{Tab:ablation_repr}.
We indicate whether the robot can achieve success in the allotted 5 epochs (5 episodes each), as well as the percentage of the task that the robot completes (Table~\ref{Tab:ablation_repr}, last column), measured by a human.

Our results in Table~\ref{Tab:ablation_repr} indicate that neither \textit{FM + $N\times$6-DoF} nor \textit{Screw + $N\times$6-DoF} representations enable task success, as  exploring in the $N\times$6-DoF space is much more challenging. 
We hypothesize that, for \textit{FM + $N\times$6-DoF}, the failing behavior is not only caused by the large uncorrelated exploration space but also by a more noisy initial trajectory that keeps the inherent noise present in the hand-tracking module.
In contrast, the use of screw actions enables \methodname{} to clean the perceived human demonstration and also makes the fine-tuning process more efficient by exploring in the reduced screw axis space.

\begin{figure}[t]
    \centering
    \includegraphics[width= 0.98\linewidth]{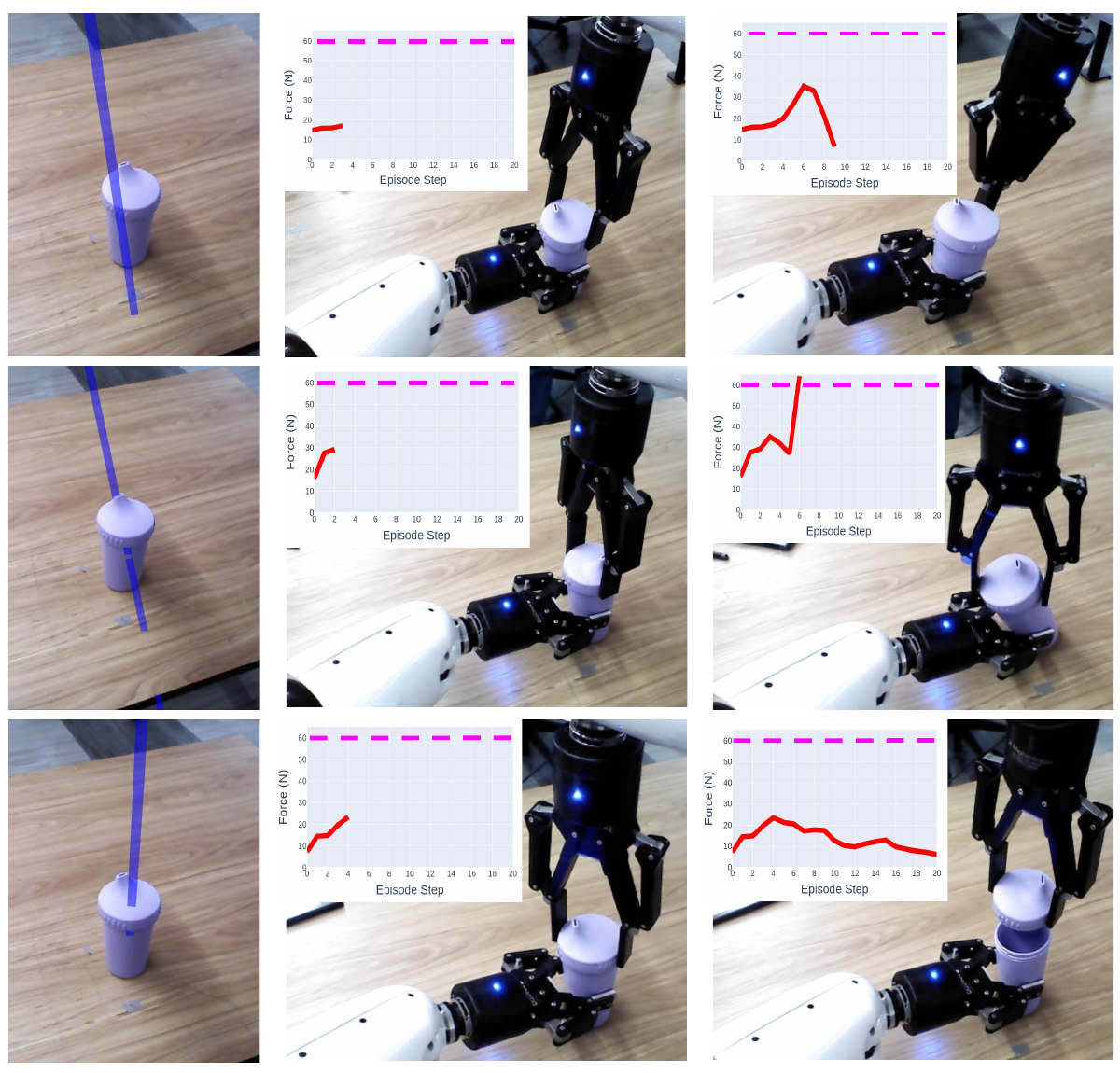}
\vspace{-3mm}
    \caption{\textbf{Reward Intuition.} Each row shows the screw axis and the snapshots of the corresponding episode to showcase the use of the reward components -- Grasp lost (first row), FT sensor reaching the threshold (second row). Each snapshot has the corresponding FT sensor reading until that timestep. The pink line shows the FT threshold. The last row shows an example of a success with the corresponding FT sensor readings. }  
    \vspace{-1em}
    \label{fig:results_reward}
\end{figure}

\vspace{2mm}
\textit{Q4) Are both autonomous reward signals correctly guiding the policy fine-tuning stage?}
The screw action policy fine-tuning stage requires a way to rank episodes in our CEM procedure, 
guiding the robot to explore around good episodes while disregarding bad ones to converge to success.
\methodname{} uses two signals to rank any episode as described in Sec.~\ref{sec:fine-tuning}: the length of an episode (based on a loss of grasp/fixation or exceeding a Force-Torque (FT) threshold), and the mean wrench measured over the episode. 
To assess the importance of these two signals for the fine-tuning stage, we ablate each component individually. 
Since removing the FT sensor threshold can be dangerous for the robot and the objects, we always retain it. 
Fig.~\ref{fig:results_reward} provides examples of the roles of these reward terms.
The results in Table~\ref{Tab:reward_abl} indicate that for the \texttt{open bottle} and \texttt{stir} tasks, if either of the two components is absent, the policy fine-tuning process fails and the robot fails to complete the task within the allotted rollouts. 
For the \texttt{insert roll} task, there was never a grasp lost in any episode, so \methodname{} succeeds even without the grasp loss detection. However, it fails without the mean episode sensed wrench. This shows that both reward signals are critical for \methodname{}'s policy fine-tuning stage.

\begin{table}[t]
\caption{Ablation of Reward Components} 
\centering
\begin{tabular}{ll|ccc} 
    \toprule
     & \multirow{2}{*}{Task} & \multicolumn{3}{c}{} \\ 
     & & Success? & \#Episodes & Dense Metric \\
    \midrule
    w/o Grasp Lost & Bottle & No & 25/25 & 20\%\\
     Detection & Roll & Yes & 7/25 & 100\% \\
     & Stir  & No & 25/25 & None\\
    \midrule
    w/o Mean Episode & Bottle & No & 25/25 & 30\%\\
     FT & Roll & No & 25/25 & 10\%\\
     & Stir & No & 25/25 & None\\
     \midrule
     \methodname{} & Bottle & Yes & 16/25 & 100\%\\
     & Roll & Yes & 7/25 & 100\%\\
     & Stir & Yes & 18/25 & None\\
    \bottomrule  
\end{tabular}
\label{Tab:reward_abl}
\vspace{-3mm}
\end{table}

\section{Lessons and Conclusion}
\label{s_conc}


In this work, we present and validate \methodname{}, a robust representational framework for bimanual manipulation that significantly boosts performance by simplifying complex tasks into screw actions derived from a single human demonstration. 
While our results demonstrate the capabilities of \methodname{}, it is not without limitations and scope for future work. 
First, the screw action formulation, although versatile, does not fit all bimanual manipulation tasks, e.g., tasks where the hands are not constrained to move along a single axis, such as cutting in a zig-zag motion. Future work can extend \methodname{} to include sequences of screw axes, requiring more complex inference and exploration algorithms. 
Second, we train a separate prediction model for each object class, which limits the generalization capabilities of \methodname{}. This can be addressed by training a single multi-task model on a diverse array of objects.
Large-scale human-activity datasets~\cite{grauman2022ego4d,Damen2018EPICKITCHENS} offer an exciting avenue to scale up the range of tasks and objects that \methodname{} can learn from. For that, our method should also relax the dependency on depth sensors, which could be obtained instead from RGB using recent algorithms~\cite{birkl2023midasv3,yang2024depthanything}.
Third, episode success is recorded manually in our experiments. This could be automated in the future using vision-language foundation models~\cite{radford2021clip,alayrac2022flamingo}. 
Fourth, while \methodname{} focuses on improving the screw axis prediction, it could be beneficial to also fine-tune the grasp contact points.
Finally, due to 3D sensor limitations, some reflective surfaces such as the laptop cannot be correctly perceived from all angles and we need to cover them. We do not deem this a problem of \methodname{} but rather of the (relatively outdated) depth sensor---using more modern 3D sensors would alleviate it.
Despite these limitations, \methodname{} demonstrates that using a screw axis space representation for bimanual actions facilitates efficient exploration leading to strong improvements in task success. Additionally, the incorporation of a self-supervised fine-tuning process allows the robot to iteratively refine its own actions. Our work is a promising step towards enabling robots to efficiently learn complex bimanual manipulation tasks by watching humans. 



\bibliographystyle{unsrtnat}
\bibliography{references}

\clearpage
\newpage
\appendix

\subsection{Screw Action with Left-Hand Trajectory}
\label{sec:left_tau}

\begin{figure}[h]
    \centering
\includegraphics[width=0.31\columnwidth]{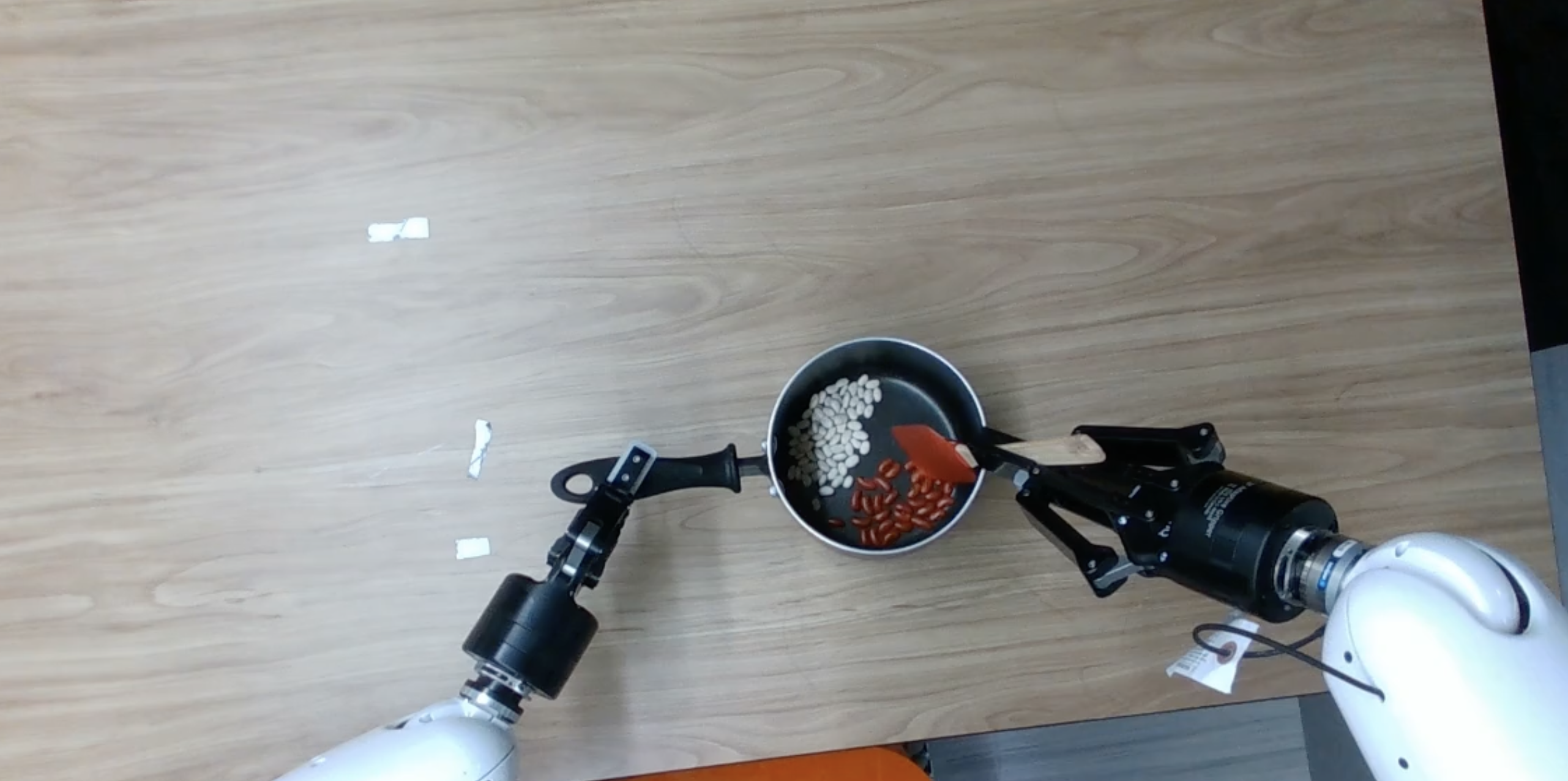}%
\hfill
\includegraphics[width=0.31\columnwidth]{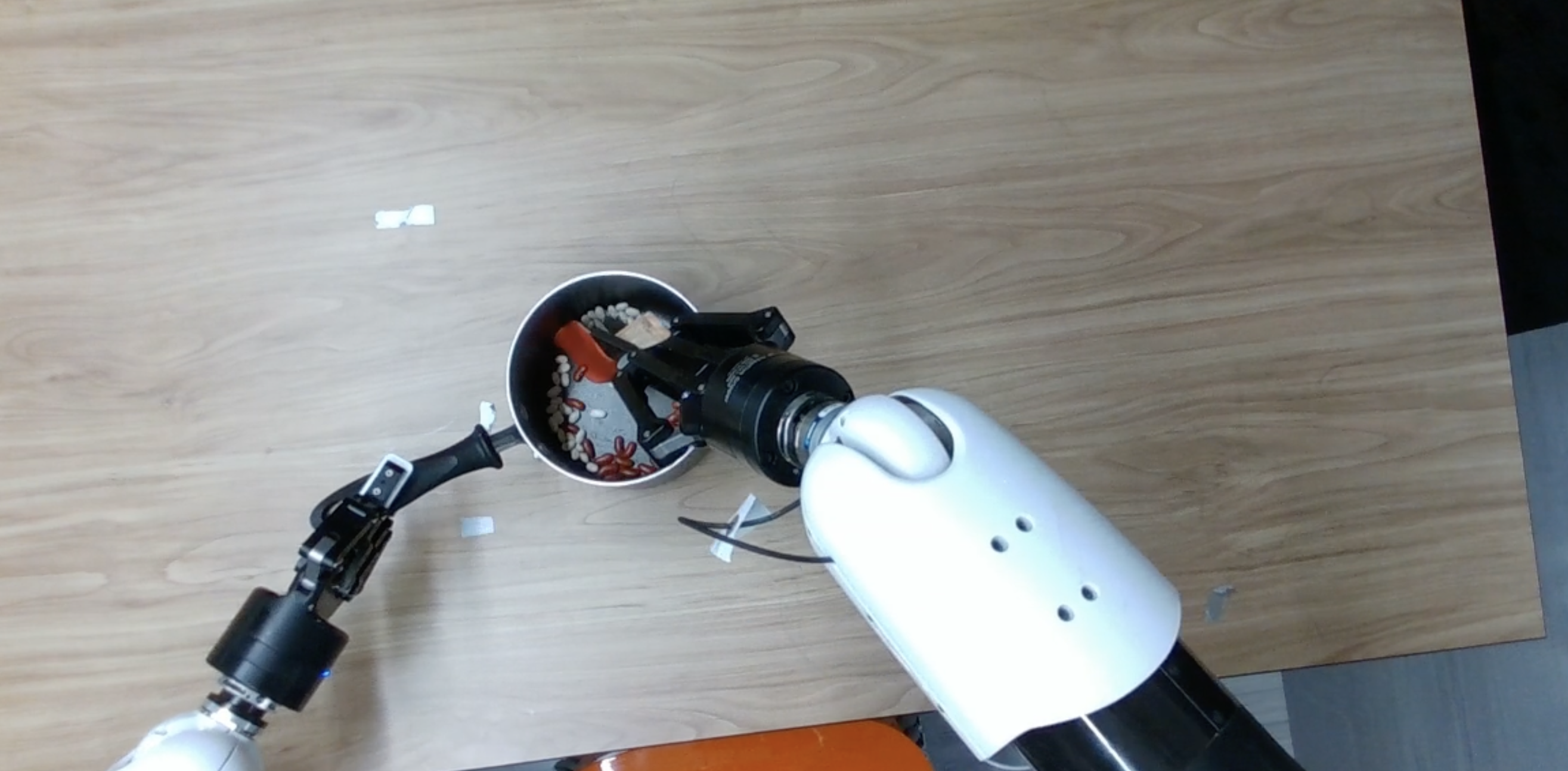}%
\hfill
\includegraphics[width=0.31\columnwidth]{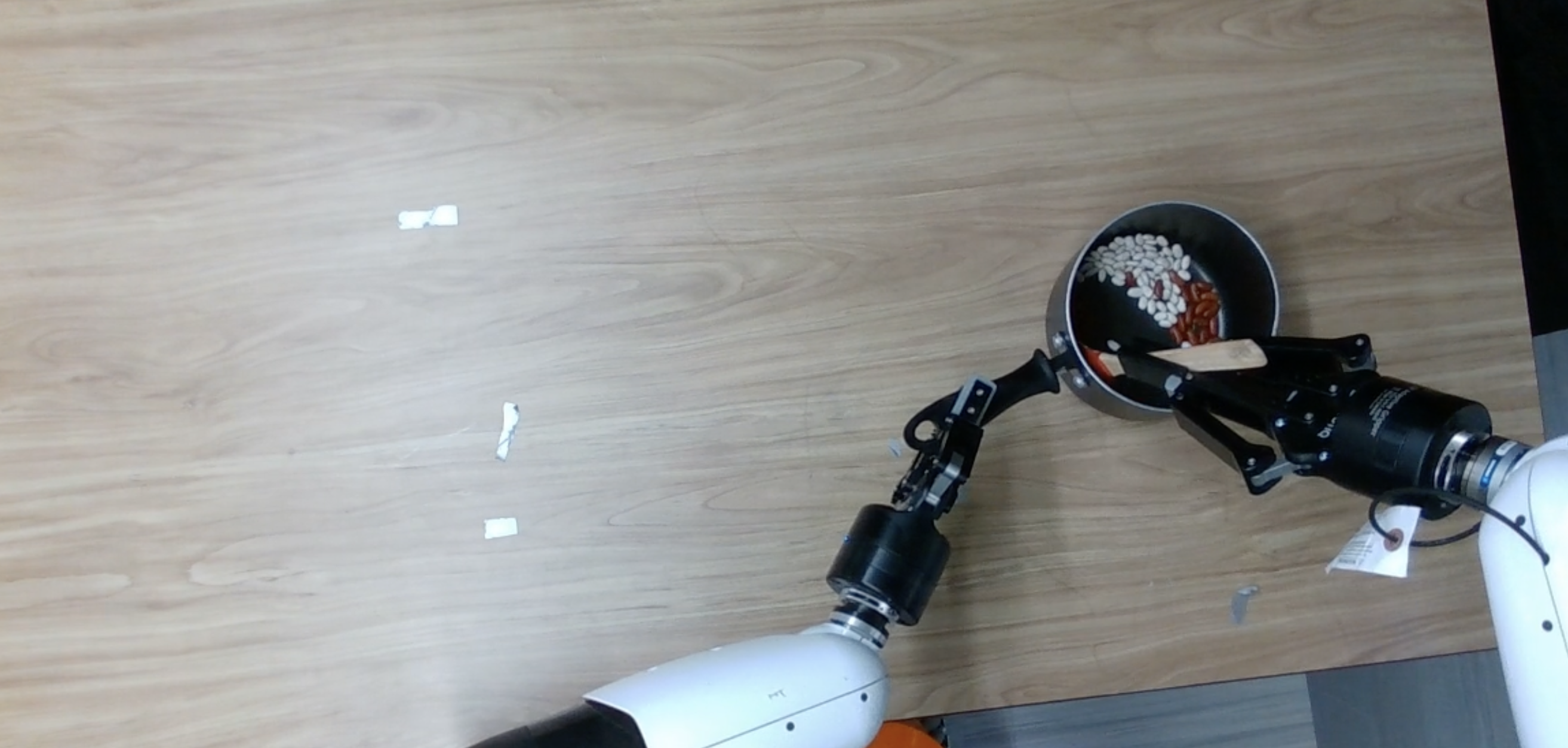}%
    \caption{\textbf{Screw Action execution with left-hand trajectory} Three steps of a robot execution of a \texttt{stir} task with non-zero left-hand motion. Since \methodname{} focuses on the generation of relative motion between hands, the same screw action can be used even when there exists any absolute motion of one of them (left hand).}
    \label{fig:lh}
\end{figure}

Fig.~\ref{fig:lh} depicts three steps of a robot execution with a non-empty left-hand trajectory. Since ScrewMimic defines the screw axis of manipulation as the relative motion between both hands, the absolute motion of one of them does not affect the motion generated from the same screw action. We consider the \textit{actuation} part of the bimanual manipulation to be the effect of this relative motion between hands rather than the absolute motion of them.

\subsection{Hyperparameters}
\begin{table}[h]\centering
\caption{Screw Action Prediction Model Hyperparameters}
\begin{tabular}{|l|c|}
    \hline
    \textbf{Hyperparameters} & \textbf{Value} \\
    \hline
    train epochs & 2000\\
    batch size & 16\\
    optimizer & Adam\\
    learning rate & 1e-3\\
    pointcloud encoder & PointNet~\cite{qi2016pointnet}\\
    number of points & 2048 \\
    layer-activations & ReLU\\
    \hline
    \multicolumn{2}{|c|}{\textbf{Screw Axis Regression}}\\
    \hline    
    Architecture & Conv1d (64, 128, 1024) \\& + FC (1024, 512, 256, 6)\\
    Loss & MSE \\
    \hline
    \multicolumn{2}{|c|}{\textbf{Grasp Contact Segmentation}}\\
    \hline
    Architecture & Conv1d (64, 128, 128, 512, 2048, \\& \qquad\qquad 256, 256, 128, 3) \\
    Loss & negative log likelihood\\

    \hline
\end{tabular}
\label{tab:hyperparam}
\end{table}

\subsection{Using Pretrained PointNet Model}
\label{sec:pretrained_pointnet}
\begin{figure}[h]
    \centering
    \includegraphics[width= 0.97\linewidth]{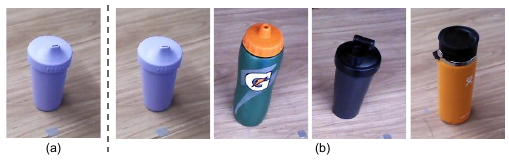}
    \vspace{-2mm}
    \caption{(a) Training bottle. (b) Testing Bottles} 
    \label{fig:bottles}
\end{figure}

We conduct experiments to analyze if using a pre-trained PointNet model helps to better generalize to new objects as compared to training from scratch. We pretrain a PointNet model on the ModelNet-40 dataset \cite{modelnet} for the classification task. We then use the pretrained feature encoder and fine-tune it on the screw axis prediction task. Finally, we compare this model (\textit{M2}) to our original model that was trained from scratch (\textit{M1}) on the screw action prediction task. The test set consists of 4 different bottles (1 bottle seen during training and 3 unseen bottles) as shown in Fig.~\ref{fig:bottles}
The test set consists of 10 poses for each of the four bottles. We use two metrics to evaluate their performance:

\begin{itemize}
    \item Mean distance between predicted and ground truth screw axis (in meters)
    \item Mean angle between the predicted and ground truth screw axis (in degrees)
\end{itemize}

As shown in Table \ref{Tab:pretrained_exp}, with a pre-trained PointNet, we indeed observe an improvement in the screw axis prediction on novel objects, although the performance on the training object remains the same. While this would not affect the exploration in case of the training object, it would lead to more efficient optimization in the CEM phase for novel objects and better generalization capabilities of the policy

\begin{table}[h] 
    \begin{center}
        \caption{PointNet trained from scratch (M1) vs Pretrained+Finetuned (M2)} \vspace*{0.5em}
        \begin{tabular}{ l|cc }
            \toprule            
              & M1 & M2         \\
            \midrule
             Blue Bottle (Training bottle) & 0.01 m, 2.1\degree & 0.01 m, 2.0\degree\\
             Green Bottle (with orange cap) & 0.03 m, 3.6\degree & 0.01 m, 2.3\degree \\
             Black Bottle & 0.05 m, 5.3\degree & 0.02 m, 3.8\degree \\
             Orange Bottle (with black cap) & 0.09 m, 6\degree & 0.03 m, 4.2\degree \\
            \bottomrule
        \end{tabular}
        \label{Tab:pretrained_exp}
    \end{center}
    \vspace{-2em}
\end{table}

\subsection{Robustness to Noisy Demonstrations}
\label{sec:noisy_demo}
To analyze the robustness of \methodname{} to noisy human demonstrations we conduct the following two experiments:

\begin{figure*}[t!]
    \centering
    \includegraphics[width= 1\textwidth]{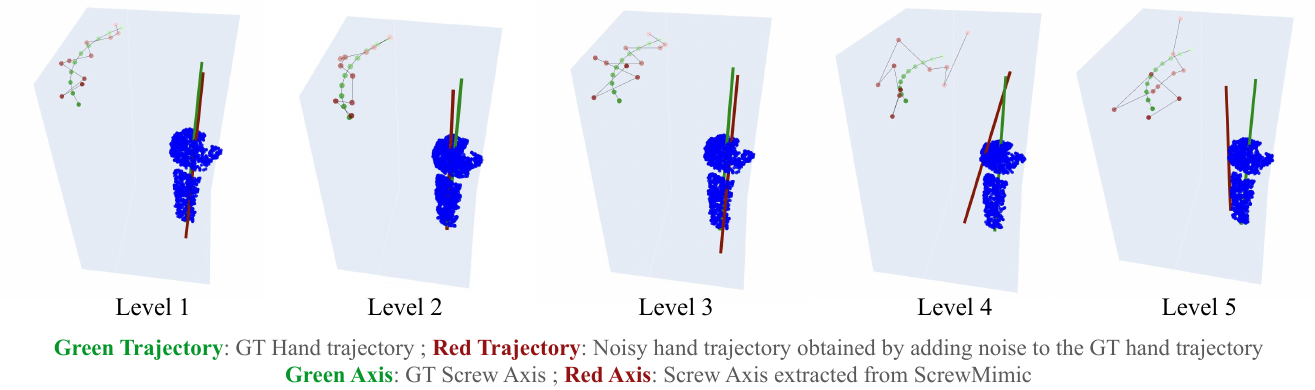}
    \vspace{-2mm}
    \caption{\textbf{ScrewMimic's axis extraction with increasingly noisy demonstrations.} 
    The five levels represent increasing noise applied to the ground truth hand trajectory (position and orientation). Comparing each screw axis with the ground truth screw axis in this figure and the numbers in Tab.~\ref{Tab:noise_exp_1}, and performing the fine-tuning experiment with the axis inferred from the highest noise level, we observe that although ScrewMimic does suffer from increasing noise in the hand trajectories, it is able to extract an axis sufficiently accurate for fine-tuning.}
    \label{fig:noise_exp_1}
\end{figure*}

\begin{table*}[t!] 
\begin{center}
    \caption{ScrewMimic’s Axis Extraction with Increasingly Noisy Demonstrations} 
    \vspace*{-0.5em}
    \resizebox{\textwidth}{!}{%
    \begin{tabular}{ l|ccccc }
        \toprule            
          & \makecell{Level 1 \\ pos=$N(0, 1.0cm)$ \\ orn=$N(0, 2.5\degree)$} 
          & \makecell{Level 2 \\ pos=$N(0, 1.5cm)$ \\ orn=$N(0, 5.0\degree)$}  
          & \makecell{Level 3 \\ pos=$N(0, 2.0cm)$ \\ orn=$N(0, 7.5\degree)$}  
          & \makecell{Level 4 \\ pos=$N(0, 2.5cm)$ \\ orn=$N(0, 10.0\degree)$}  
          & \makecell{Level 5 \\ pos=$N(0, 3.0cm)$ \\ orn=$N(0, 12.5\degree)$}         \\
        \midrule
          Distance between \\ GT axis and \\ Extracted axis (cm) 
         & $0.5 cm \pm 10^{-5}$
         & $0.8 cm \pm 10^{-4}$
         & $1.1 cm \pm 10^{-5}$
         & $1.7 cm \pm 10^{-4}$ 
         & $2.1 cm \pm 10^{-2}$\\
         \midrule
         Angle between \\ GT axis and 
         \\ Extracted axis (degrees) 
         & $4.0\degree \pm 2.0\degree$
         & $6.5\degree \pm 3.0\degree$ 
         & $9.4\degree \pm 5.0\degree$ 
         & $11.1\degree \pm 8.5\degree$
         & $13.1\degree \pm 6.0\degree$ \\
        \bottomrule
    \end{tabular}
    }
    \label{Tab:noise_exp_1}
\end{center}
\vspace*{-1.5em}
\end{table*}

\paragraph{\textbf{Artificially adding increasing amounts of noise (controlled study)}} 
In the first experiment, we investigate how well ScrewMimic can adapt when increasing amounts of artificial noise are introduced to a trajectory.  We focus on a specific task — \texttt{open bottle}. We manually annotate the ground truth screw axis and compute the corresponding noise-free ground truth hand trajectory (shown in green in Fig. \ref{fig:noise_exp_1}). 
This trajectory corresponds to the trajectory of an acting hand relative to a reference hand.
We introduce five different levels of noise to the ground truth hand trajectory, affecting both position and orientation, and observe the changes in the screw axis computed by ScrewMimic with increasing noise levels. These trajectories and their respective screw axes are illustrated in Fig.~\ref{fig:noise_exp_1}, where colors transitioning from light to dark depict the sequence of actions from start to finish. We only visualize the positions (and not the orientations) for clarity. We created 20 noisy trajectories for each noise level, resulting in 100 test trajectories. We use two metrics to evaluate the performance for each noise level: a) mean distance error between predicted and ground truth screw axes (in meters), and b) mean angle error between the predicted and ground truth screw axis (in degrees).

Tab.~\ref{Tab:noise_exp_1} and Fig.~\ref{fig:noise_exp_1} show the results of our experiment. 
We observe that the accuracy of the screw axis detected by ScrewMimic declines as we increase the noise in the hand trajectories. 
To test if ScrewMimic can perform a successful fine-tuning even with the highest noise level, we conduct the following experiment:
we use the axis inferred by ScrewMimic from the trajectory in level 5 to bootstrap the ScrewMimic fine-tuning step. We observe that even in this adversarial condition, success is achieved after 4 epochs and 21 episodes. This is comparable to the performance of ScrewMimic on the bottle opening task in our original experiments as shown in Tab. \ref{Tab:main}. 
Thus, ScrewMimic is able to “clean up” the noise and extract an axis good enough to bootstrap the fine-tuning step.
This shows that even though the quality of the screw axis inferred by ScrewMimic declines with increasing noise, the axis still proves adequate for initiating the fine-tuning process.

\begin{figure*}[t!]
    \centering
    \includegraphics[width=1\textwidth]{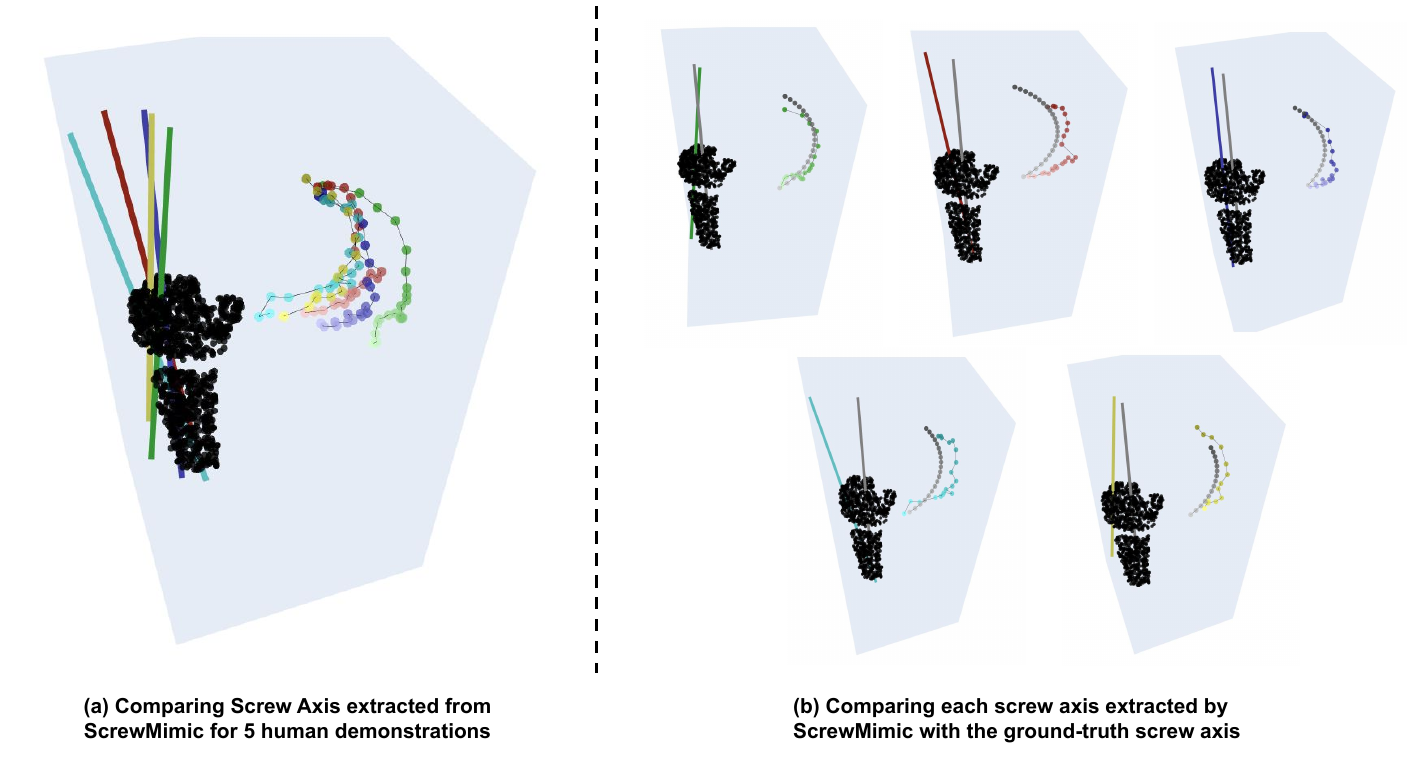}
    \vspace{-2mm}
    \caption{\textbf{Screw axis extracted for five human demonstrations.} (a) Five human trajectories and their corresponding screw axes extracted by ScrewMimic. (b) Individual trajectories and extracted screw axis along with the ground truth trajectory and screw axis. Despite the diversity in the trajectories due to variations in demonstrations and detection noises, ScrewMimic is able to extract a screw axis sufficiently accurate for fine-tuning.} 
    \label{fig:noise_exp_2}
\end{figure*}

\paragraph{\textbf{Naturally occurring noise (perceptual noise)}} In the second experiment, we evaluate the robustness of ScrewMimic to naturally occurring noise when perceiving human demonstrations. We collect five different human demonstrations for the bottle opening task for the same pose of the bottle. Variability in the trajectories arises from differences in individual demonstrations and noise from the hand-pose detector. We compare the screw axis computed by ScrewMimic for these five demonstrations to a manually annotated ground truth axis. Fig.~\ref{fig:noise_exp_2} shows the qualitative results for this experiment: Fig.~\ref{fig:noise_exp_2} (a) on the left helps visually compare the five human trajectories and the corresponding screw axes as computed by ScrewMimic. Note that the trajectory corresponds to the trajectory of the acting hand (right hand in this case) relative to the reference hand (left hand). 
Fig.~\ref{fig:noise_exp_2} (b) on the right shows a comparison of the trajectory and computed screw axis with the ground truth trajectory and screw axis for each of the five human demonstrations.

Tab.~\ref{Tab:noise_exp_2} shows the quantitative results of the distance error and angle error between the axis computed by Screwmimic and the ground truth axis. 
These axis errors are comparable to the axis errors for the \texttt{open bottle} task in our original experiments (refer to Sec. \ref{s_exp} Q1) which are $1.42 cm$ mean distance error and $11.2 \degree$ mean angle error. 
As can be seen in the first row of Tab. \ref{Tab:main} in the main paper, ScrewMimic's fine-tuning process is able to correct the initial noisy axis and succeed at the task for the most part. 
Since the errors in the axes as shown in Tab. \ref{Tab:noise_exp_2} are comparable to the error in our original \texttt{open bottle} experiments, we can infer that ScrewMimic would be able to fine-tune these noisy axes.
This shows that despite the diversity in the trajectories due to variations in demonstrations and detection noise, ScrewMimic consistently infers a screw axis with an accuracy that proves adequate for initiating the fine-tuning process.

\begin{table*}[t!] 
\begin{center}
    \caption{Analyzing Screw Axis Extracted from Five Human Demonstrations} 
    \vspace*{0.5em}
    \resizebox{\textwidth}{!}{%
    \begin{tabular}{ l|ccccc }
        \toprule            
          & \textcolor[HTML]{049629}{Demo 1} & \textcolor[HTML]{941213}{Demo 2} & \textcolor[HTML]{4041aa}{Demo 3} & \textcolor[HTML]{4ac0c2}{Demo 4} & \textcolor[HTML]{bebf4e}{Demo 5}        \\
        \midrule
         Distance between GT and \\ Extracted axes (cm) & 0.91 cm 
         & 1.26 cm & 1.45 cm & 1.33 cm & 1.10 cm\\
         \midrule
         Angle between GT axis \\ and Extracted axis (degrees) & 6.0\degree &  
         6.5\degree & 6.4\degree & 12.3\degree & 8.7\degree\\
        \bottomrule
    \end{tabular}
    }
    \label{Tab:noise_exp_2}
\end{center}
\end{table*}

\end{document}